\documentclass[final,12pt]{elsarticle}
\usepackage[]{natbib}
\usepackage{algorithmic}
\usepackage{amsfonts}
\usepackage{amsmath}
\usepackage{amssymb}
\usepackage{array}
\usepackage{balance}
\usepackage{bbding}
\usepackage{booktabs}
\usepackage{cite}
\usepackage{color}
\usepackage{graphicx}
\usepackage{hyperref}
\usepackage{lineno}
\usepackage{makecell}
\usepackage{multicol}
\usepackage{multirow}
\usepackage{stfloats}
\usepackage{textcomp}
\usepackage{verbatim}
\usepackage{url}

\journal{arXiv}

\begin{document}
\begin{frontmatter}

\title{AoSRNet: All-in-One Scene Recovery Networks via Multi-knowledge Integration}
\author[1]{Yuxu Lu}
\ead{yuxulouis.lu@connect.polyu.hk}
\author[1]{Dong Yang}
\ead{dong.yang@polyu.edu.hk} 
\author[2]{Yuan Gao}
\ead{yuangao@whut.edu.cn} 
\author[2]{Ryan Wen Liu}
\ead{wenliu@whut.edu.cn} 
\author[3]{Jun Liu}
\ead{liuj292@nenu.edu.cn} 
\author[2]{Yu Guo}
\ead{yuguo@whut.edu.cn} 
\affiliation[1]{organization={Department of Logistics and Maritime Studies, The Hong Kong Polytechnic University},
	postcode={999077}, 
	state={Hong Kong}}
\affiliation[2]{organization={School of Navigation, Wuhan University of Technology},
	city={Wuhan},
	postcode={430063}, 
	state={China}}
%
%
\affiliation[3]{organization={School of Mathematics and Statistics, Northeast Normal University},
	city={Changchun},
	postcode={130024}, 
	state={China}}
\begin{abstract}
    Scattering and attenuation of light in no-homogeneous imaging media or inconsistent light intensity will cause insufficient contrast and color distortion in the collected images, which limits the developments such as vision-driven smart urban, autonomous vehicles, and intelligent robots. In this paper, we propose an all-in-one scene recovery network via multi-knowledge integration (termed AoSRNet) to improve the visibility of imaging devices in typical low-visibility imaging scenes (e.g., haze, sand dust, and low light). It combines gamma correction (GC) and optimized linear stretching (OLS) to create the detail enhancement module (DEM) and color restoration module (CRM). Additionally, we suggest a multi-receptive field extraction module (MEM) to attenuate the loss of image texture details caused by GC nonlinear and OLS linear transformations. Finally, we refine the coarse features generated by DEM, CRM, and MEM through Encoder-Decoder to generate the final restored image. Comprehensive experimental results demonstrate the effectiveness and stability of AoSRNet compared to other state-of-the-art methods. The source code is available at \url{https://github.com/LouisYuxuLu/AoSRNet}.
\end{abstract}


\begin{highlights}
    \item 	AoSRNet improves imaging performance in hazy, sandy, and low-light degraded scenes.

    \item 	Multi-knowledge integration strategy robustly restores image in unpredictable scenes.

    \item 	Optimized linear stretching and gamma correction improve AoSRNet's generalization.

    \item 	We constructed a set of atmospheric light values for haze and sand image synthesis.

    \item 	Extensive experiments verify AoSRNet's effectiveness with start-of-the-art methods.
\end{highlights}


\end{frontmatter}


\section{Introduction}
\begin{figure}[tb]
        \centering
        \setlength{\abovecaptionskip}{0.cm}
        \includegraphics[width=0.75\linewidth]{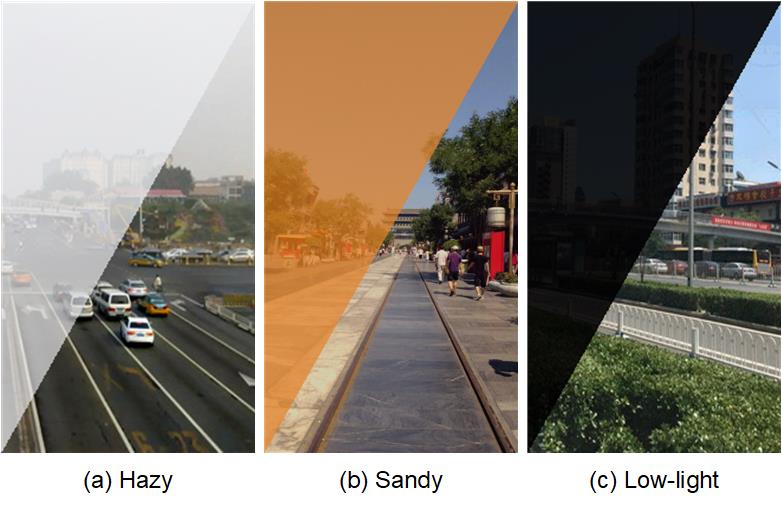}
        \caption{Example of the scene recovery in three different imaging conditions. The upper triangles in (a)-(c) are degraded patterns, and the corresponding restored patterns by our method are shown in the lower triangles.}
        \vspace{-0.2cm}
        \label{Figure01}
    \end{figure}

    Imaging quality is typically impacted by unpredictable degradation factors in adverse imaging environments, such as light scattering and attenuation in inhomogeneous media (e.g., haze and sand dust), local or global insufficient luminance due to inconsistent light intensity (e.g., low-lightness), etc \citep{liu2022rank}. As shown in Fig. \ref{Figure01}, the unexpected imaging environment reduces the contrast and eliminates a substantial amount of color and edge texture information. Degraded images can lead to a loss of accuracy for sophisticated and advanced vision-driven intelligent devices or systems tasks such as detection or semantic segmentation \citep{chen2023mask,chen2023edge}. Researchers have focused on extracting latent feature information from degraded imaging scenes in recent years and conducted extensive research on various degradation factor categories. Specifically, scene recovery methods are categorized mainly as traditional- or learning-based.  
    The traditional scene recovery methods are mainly composed of dark channel prior (DCP)- \citep{he2010single} and Retinex-based \citep{land1977retinex} methods. DCP can generate haze-free images without the need for accurate physical modelling of the haze, which has reasonable practicability and robustness. DCP has also been applied to other degradation scenes (e.g. sand dust \citep{kim2019fast,liu2022rank} and low light \citep{jiang2013night}). The Retinex theory \citep{land1977retinex} and its improved methods \citep{fu2014retinex,kandhway2023adaptive} obtain the illumination image by filtering the degraded image and finally separate the illumination image from the original image to generate the latent reflection image. Nonetheless, the Retinex-based methods require to be improved in complex imaging degradation scenes, such as inconsistent illumination. Gamma correction (GC) \citep{zhu2020novel} and linear stretching (LS) \citep{kim2019fast} are usually used to optimize the luminance and contrast of the image restored by the DCP- or Retinex-based methods \citep{zhang2022underwater, liu2021joint} so that the restored image has a more natural visual performance.
    Learning-based methods can recover images from degraded scenes effectively due to their powerful feature generalization capability. Some learning-based methods have achieved excellent performance in the field of image restoration, such as convolutional neural networks (CNN) \citep{jia2021effective, li2022single}, generative adversarial networks (GAN) \citep{ma2021learning,dai2024understanding}, Transformer \citep{wang2023uscformer,zhou2023physical}, and denoising diffusion probabilistic models (DDPM) \citep{croitoru2023diffusion}. However, relying solely on end-to-end mapping through deep networks can potentially lead to overfitting and limited generalization ability in various scenes. Consequently, learning methods that incorporate physical prior models (such as DCP- \citep{ren2016single} and Retinex-guided \citep{zhang2021beyond}) have been proposed. The model-driven learning methods benefit from the constraints of prior features on the learning parameters, enabling them to be more applicable to diverse scenes and enhancing the generalization ability and robustness of deep networks.
    In this work, we propose an all-in-one scene recovery network via multi-knowledge integration (termed AoSRNet) to improve the visibility of imaging devices in typical low-visibility imaging scenes (i.e., haze, sand dust, and low light). Specifically, we combine the gamma correction (GC) and optimized linear stretching (OLS) with the standard residual block (SRB), thus proposing the detail enhancement module (DEM) and color restoration module (CRM) to guide the sub-learning networks to restore the degraded image. Moreover, we suggest a multi-receptive field (MRF) extraction module (MEM) to attenuate the loss of image texture details resulting from GC nonlinear and OLS linear transformations. DEM and CRM will alleviate the overfitting of the deep network so that AoSRNet can improve the imaging quality of the visual sensor more robustly and efficiently in different degradation scenes. The Encoder-Decoder-based fusion module (EDFM) is then used to refine and fuse the coarse features generated by DEM, CRM, and MEM to generate the final image restoration. The main contributions of this work can be summarized as follows

    \begin{itemize}
        \item 	We propose an all-in-one scene recovery network (AoSRNet) via multi-knowledge integration to improve imaging performance in various degraded scenes (e.g., haze, sand dust, and low light).

        \item 	We propose a multi-knowledge (including GC-guided DEM, OLS-guided CRM, and MEM) integration strategy to achieve more robust image restoration in unpredictable degraded scenes.

        \item 	Without loss of generalization, we conduct extensive experiments to verify the effectiveness of AoSRNet in three scene recovery tasks with competitive methods.
    \end{itemize}

    The rest of this paper is organized as follows. Section \ref{sec:aosrnet} introduces the proposed AoSRNet. Numerous experiments have been implemented to evaluate the performance of AoSRNet in Section \ref{sec:experiments}. Conclusions are given in Section \ref{sec:conclusions}.	
    \begin{figure*}[t]
        \centering
        \setlength{\abovecaptionskip}{0.cm}
        \includegraphics[width=1.00\linewidth]{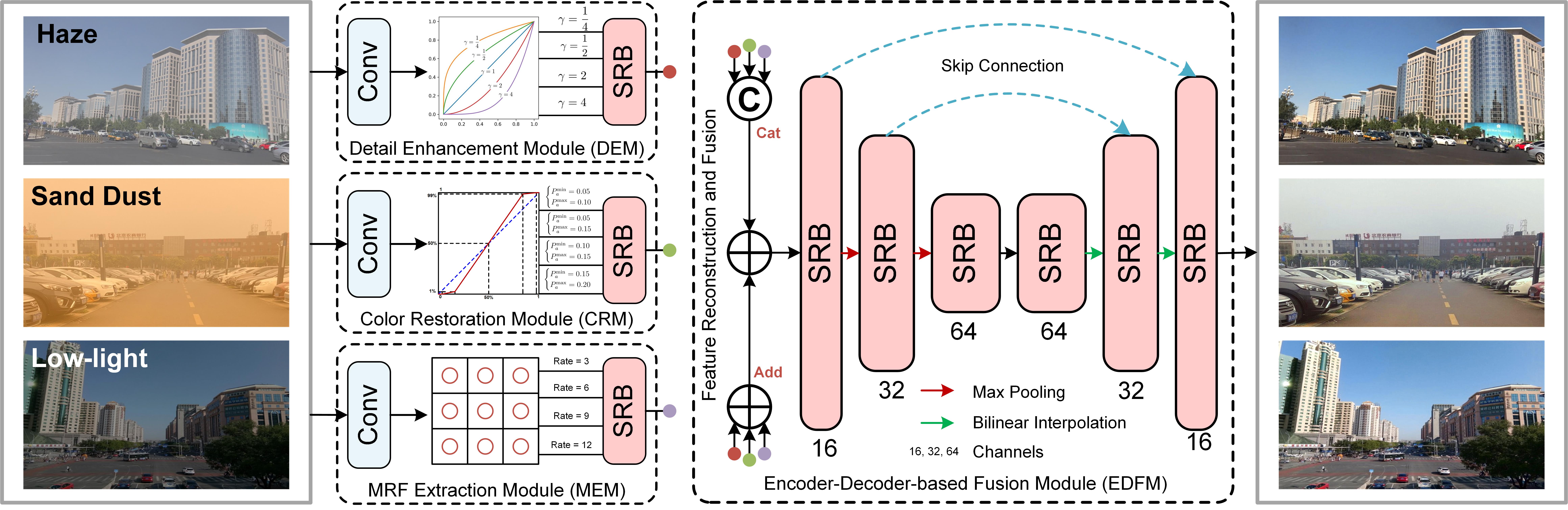}
        \caption{The flowchart of the all-in-one scene recovery network (AoSRNet). It mainly contains gamma correction (GC)-guided detail enhancement module (DEM), optimized linear stretching (OLS)-guided color restoration module (CRM), multi-receptive field (MRF) extraction module (MEM), and Encoder-Decoder-based fusion module (EDFM). Standard residual block (SRB) is the basic unit of learning.}
        \vspace{-0.3cm}
        \label{Figure_Flowchart}
    \end{figure*}
\section{Related Work}
\label{sec:relatedwork}
    A clear and explicit scene can satisfy the visual needs of humans and facilitate higher-level processing. This section concisely summarizes related works conducted under various imaging conditions.
\subsection{Dehazing}
    Image dehazing methods are generally classified into physical model- \citep{he2010single,kim2019fast,liu2022rank}  and learning-based \citep{ren2016single,zhao2021refinednet,zhou2022fsad,song2023vision,guo2023scanet}. The classic physical model-methods, that is, dark channel prior (DCP) \citep{he2010single}, generate haze-free images by revealing the statistical laws of hazy images and inverting the atmospheric scattering model. However, since the DCP-based methods are not fully applicable in the bright areas of the image (such as the sky and water surface), the image after dehazing is visually locally distorted \citep{shu2019variational}. With the application of deep learning in the field of low-level vision, end-to-end and physical model-based learning methods have been proposed. End-to-end methods (such as FFANet \citep{qin2020ffa} and FSADNet \citep{zhou2022fsad}) directly model the mapping from haze to haze-free images, reducing the need for manual feature extraction and focusing on the training data itself. However, the end-to-end method is vulnerable to the influence of training data, resulting in the phenomenon of network overfitting. Physical model-guided methods can embed the atmospheric scattering model into the network so that the imaging model is still followed in the restoration process of the degraded image, which can reduce the risk of overfitting \citep{liu2022deep}. For example, MSCNN \citep{ren2016single} and RefineDNet \citep{zhao2021refinednet} use the atmospheric scattering model to reconstruct haze-free images. AODNet \citep{li2017aod} reconstructs the atmospheric scattering model, reducing unnatural dehazed images generated due to inaccurate estimates of transmittance and atmospheric light values.
\subsection{Sand Dust Image Enhancement}
    The sand dust image enhancement task is comparable to the image dehazing task, but the atmospheric light value of the red channel of the dust image is substantially higher than that of the green and blue channels. The blue channel suffers from severe information loss \citep{cheng2020fast}. Hereby, The performance of the classic DCP-based methods \citep{he2010single,zhu2018haze} in the dust image enhancement task is not significant. Since linear stretching (LS) and Gamma correction (GC) can adjust the histogram distribution of each channel, the channel of information loss is compensated to restore the brightness and contrast of the image. Therefore, LS or GC will be used for preprocessing of degraded images in traditional dust image enhancement tasks. For example, Peng \textit{et al.} \citep{peng2018generalization} incorporated adaptive color correction into the image formation model (IFM) and proposed a generalized dark channel prior for single image restoration. Wang \textit{et al.} \citep{wang2021fast} proposed a color compensation- and affine transform-guided fast color balance and multipath fusion method for sand and dust image enhancement. Fu \textit{et al.} \citep{fu2014fusion} combined LS and GC to improve the visibility of dust images. In addition, to better compensate for the missing channel information, literature \citep{shi2020normalised} and \citep{gao2022color} map the image to the LAB space. Limited by paired sand/sand-free image datasets, the learning method is less researched than the traditional method, but it is still mainly driven by end-to-end \citep{gao2023let} and physical prior models \citep{si2022sand,ding2022restoration}. TOENet \citep{gao2023let} reconstructs the features between the three RGB channels through the channel correlation extraction module (CCEM) to restore the degraded image. Ding \textit{et al.} \citep{ding2022restoration} combined the GAN and DCP to improve the visibility of sand dust images through unsupervised learning.
\subsection{Low-light Image Enhancement}
    Low-light image enhancement has been widely studied and successfully applied in different fields. Traditional methods mainly include histogram equalization (HE)- \citep{pizer1987adaptive}, dehazing- \citep{jiang2013night} and Retinex-based \citep{land1977retinex} methods, etc. HE-based methods \citep{pizer1987adaptive,reza2004realization} accomplish image enhancement by uniformly distributing the histogram of the image's pixel intensity or color distribution. Dehazing-based methods \citep{jiang2013night,feng2020low} reverse a low-light image into a pseudo-haze image and then reverse the dehazed image to obtain a normally-illuminated image. However, there are evident differences between the pseudo-haze and real-world hazy images, making it difficult for the dehazing method to be fully applicable. Based on the Retinex theory, researchers have conducted extensive research and improved enhancement performance \citep{wang2013naturalness,guo2016lime,li2018structure}. In complex low-visibility situations, however, applying Retinex-based methods requires more work. Learning-based methods have been successfully applied to low-light enhancement tasks in recent years and are mainly researched from two aspects: end-to-end \citep{ren2019low,lu2022mtrbnet,xu2023low} and physical model-guided learning. The hybrid deep network proposed by Ren \textit{et al.} \citep{ren2019low} incorporates gradient features to improve the network's extraction of edge features covered by darkness. Usually, Retinex theory is used as the basic physical model to guide learning methods (such as RetinexNet \citep{wei2018deep}, KinD+ \citep{zhang2021beyond}, and CSDNet\citep{ma2021learning}). Physical model-driven learning methods tend to have more stable enhancement performance and more robust scene generalization ability.
\subsection{Multi-scene Recovery}
	The imaging models of different degraded scenes are different, and it is often difficult for traditional/learning methods to achieve multiple types of degraded image inpainting only through a single model \citep{liu2022rank}. Nevertheless, researchers still try and propose a lot of work. Traditional learning methods mainly use LS or GC for color correction, which uses the corrected features as prior features or post-optimization of the main repair model (e.g., DCP- \citep{peng2018generalization,wang2021fast, ancuti2017color} and Retinex-driven \citep{fu2014retinex}). Although LS and GC can assist in restoring the brightness and contrast of degraded images, it is difficult to adaptively adjust to different degraded scenes and different degraded degrees, and it is prone to over/underexposure and loss of detail texture. Learning-based multi-scene restoration methods have been proposed successively, such as TOENet (for haze and sand dust) \citep{gao2023let}, LYSNet (for haze and low-light) \citep{qu2023deep} and DIA (for haze and low-light) \citep{kim2021deep}. The learning methods can alleviate the problem of insufficient adaptive ability of traditional methods through learnable network parameters. However, they are also easy to overfit, which leads to the insufficient generalization ability of the network. Therefore, a key strategy for enhancing multi-scene robustness and generalization ability is to combine learning methods with traditional physical model or prior information.
\section{All-in-One Scene Recovery Network}\label{sec:aosrnet}
    Visible light imaging systems have penetrated into various fields of vision-driven intelligent devices or systems. However, undesired degraded scenes (e.g., haze, sand dust, and low light) degrade the imaging quality, thus affecting the performance of advanced vision tasks. To address this problem, as shown in Fig. \ref{Figure_Flowchart}, we propose a novel AoSRNet, which mainly contains gamma correction (GC)-guided detail enhancement module (DEM), optimized linear stretching (OLS)-guided color restoration module (CRM), multi-receptive field (MRF) extraction module (MEM), and Encoder-Decoder-based fusion module (EDFM). The proposed total loss function consists of $\ell_1$-norm loss $\mathcal{L}_{\ell_1}$, contrastive regularization loss $\mathcal{L}_{cr}$, and color loss $\mathcal{L}_{\text {color}}$. AoSRNet can achieve satisfactory performance on multi-scene recovery tasks through a single network.
\subsection{Standard Residual Block}
    \begin{figure}[tb]
        \centering
        \setlength{\abovecaptionskip}{0.cm}
        \includegraphics[width=0.75\linewidth]{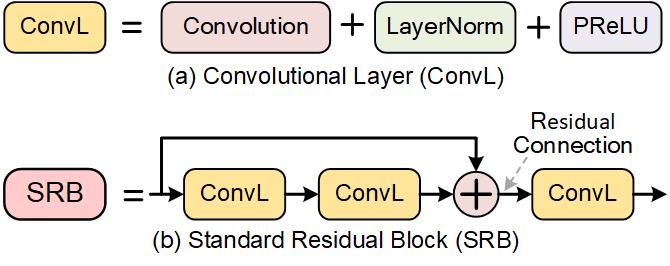}
        \caption{Suggested basic composition of (a) Convolutional Layer (ConvL) and (b) standard residual block (SRB).}
        \vspace{-0.3cm}
        \label{Figure04}
    \end{figure}
    Residual learning has demonstrated its efficient performance in different fields of computer vision. So we propose a standard residual block (SRB, $\textbf{SRB}(\cdot)$) as the basic learning unit of the proposed network. As shown in Fig. \ref{Figure04}, a SRB consists of three convolutional layers (ConvL, $\textbf{ConvL}(\cdot)$), and a ConvL sequentially contains convolutional operation ($\mathcal{C}(\cdot)$), layer normalization (LayerNorm, $\mathcal{N}(\cdot)$), and parametric rectified linear unit (PReLU, $\mathcal{P}(\cdot)$), which can be given as
    \begin{equation}\label{eq:convolutional}
        \textbf{ConvL}(x_{cl})=\mathcal{P}(\mathcal{N}(\mathcal{C}(x_{cl}))),
    \end{equation}
    where $x_{cl}$ is the input of the $\textbf{ConvL}(\cdot)$. LayerNorm can better balance the extraction of single-layer channel features and the association of multi-layer channel features in image restoration tasks. Therefore, the suggested SRB can be defined as
    \begin{equation}
        \textbf{SRB}(x_{in})=\mathcal{P}(\mathcal{N}(\mathcal{C}(\textbf{ConvL}(\textbf{ConvL}(x_{in})))+x_{in}),
    \end{equation}
    where $x_{in}$ is the input of the $\textbf{SRB}(\cdot)$. As the basic learning module, SRB will be applied to construct the all-in-one recovery network with stable enhancement performance and low computational cost.
\subsection{Detail Enhancement Module}
    The pixel values of the single/three channels in hazy, sandy, and low-light scenes are enlarged or reduced nonlinearly, resulting in poor image contrast and unnatural visual phenomena of overexposure or underexposure. GC can perform nonlinear operations on the gray value of the degraded image. By using different $\gamma$-coefficients, the gray value of the output image is exponentially related to the gray value of the input image. GC can achieve a satisfactory balance between image texture information preservation and image exposure adjustment.
    \begin{figure}[tb]
        \centering
        \setlength{\abovecaptionskip}{0.cm}
        \includegraphics[width=0.75\linewidth]{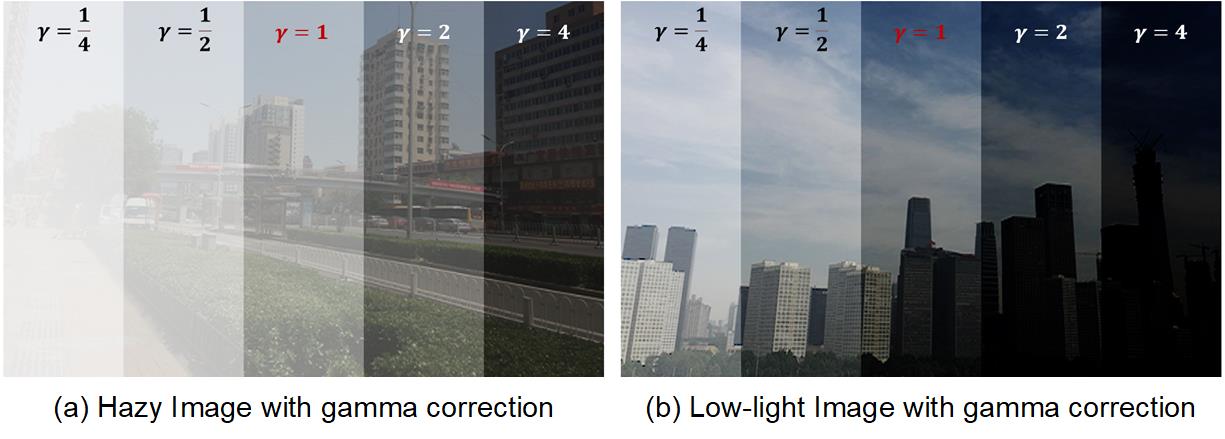}
        \caption{Processed results of gamma correction operations of (a) hazy and (b) low-light image with different $\gamma$ values (i.e., $\gamma=\frac{1}{4},\frac{1}{2},1,2$, and $4$).}
        \vspace{-0.3cm}
        \label{Figure05}
    \end{figure}
    The brightness and contrast of degraded images are uncertain and complex. For example, images captured under hazy scenes tend to have higher brightness levels, while the opposite is true for low-light scenes. As shown in Fig. \ref{Figure05}, when $\gamma<1$, GC will increase the brightness of the underexposed image, and when $\gamma>1$, the contrast of the overexposed image will be enhanced. The contrast of the image can be defined as
    \begin{equation}
        \psi(\Omega)=I_{\max }^{\Omega}-I_{\min }^{\Omega},
    \end{equation}
    where $I_{\max }^{\Omega}=\max \{I(x) \mid x \in \Omega\} \text { and } I_{\min }^{\Omega}=\min \{I(x) \mid x \in \Omega\}$. $I(x)$ is the degraded image. Therefore, to meet the correction needs of different degradation scenes and mine more detailed feature information, Gamma correction (GC, $\textbf{GC}(\cdot)$) with four different $\gamma$ values (i.e., $\gamma=\frac{1}{4},\frac{1}{2},2$, and $4$) are used to obtain the corresponding over/under-exposed images. The intensity of the globally modified image by a power function transformation can be expressed as
    \begin{equation}
        I_{\text{gc}} = \varepsilon \cdot I^\gamma =\textbf{GC}(I),\quad \gamma=\frac{1}{4},\frac{1}{2} ,2, 4,
    \end{equation}
    where $\varepsilon$ is a positive real constant, usually $\varepsilon=1$. We embed GC-guided features into SRB (i.e., detail enhancement modules, DEM, $\textbf{DEM}(\cdot)$), which can more robustly guide the network to extract valuable brightness, local texture features, etc. Therefore, the DEM can be given as
    \begin{equation}\label{eq:dem}
        \textbf{DEM}(I(x))=\textbf{SRB}(\textbf{GC}(\mathcal{C}(I(x)))).
    \end{equation}
    The proposed DEM provides valuable feature guidance for the recovery network, which can improve the network convergence speed and the generalization ability of multi-degradation scenes.
    \begin{figure}[tb]
        \centering
        \setlength{\abovecaptionskip}{0.cm}
        \includegraphics[width=0.75\linewidth]{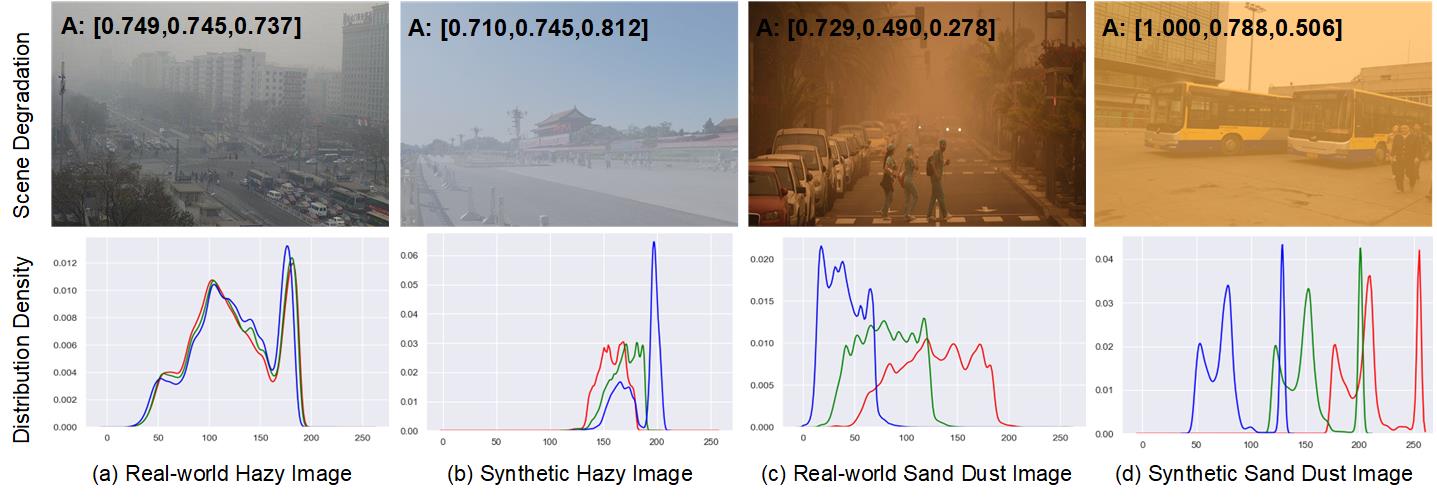}
        \caption{The atmospheric light value and its corresponding three-channel histogram distribution of the two types of degraded scenes.}
        \vspace{-0.3cm}
        \label{Figure02}
    \end{figure}
\subsection{Color Restoration Module}
    The gray pixel value of the degraded image concerned in this paper is distributed in a small range, but the gray level is large, as shown in Fig. \ref{Figure02}, which leads to low utilization of the gray level and a serious decrease in the contrast of the image. Therefore, to rescale the gray histogram distribution of degraded images, we adopt OLS to diffuse the pixel distribution in a small range to the whole gray level range. In general, LS can be defined as
    \begin{equation}\label{eq:ls}
        I_{\text{ls}}(x)=I_{\text{ls}}^{\min}+\frac{I_{\text{ls}}^{\max}-I_{\text{ls}}^{\min}}{I^{\max}-I^{\min}} \cdot\left(I(x)-I^{\min}\right),
    \end{equation}
    where $I(x)$, $I_{\text{ls}}(x)$ are the pixel values of the degraded image and the stretched image; $I^{\min}$, $I^{\max}$ are the minimum and maximum value of the degraded image pixel value; $I_{\text{ls}}^{\min}$ and $I_{\text{ls}}^{\max}$ are the minimum and maximum value of the stretched image pixel values. Generally, it is assumed that $I_{\text{ls}}^{\min} = 0$, $I_{\text{ls}}^{\max} = 1$. Therefore, Eq. \ref{eq:ls} can be redefined as
    \begin{equation}
        I_{\text{ls}}(x)=\frac{I(x)-I^{\min}} {I^{\max }-I^{\min}}.
    \end{equation}
    Different imaging environments have differences in light scattering and attenuation. Therefore, we recommend optimized linear stretching (OLS, $\textbf{OLS}(\cdot)$) to better control the intermediate tones, shadows, and highlights in the image. It calculates the minimum and maximum stretching values based on four values, i.e., minimum percentage $P^{\min}$, maximum percentage $P^{\max}$, minimum adjustment percentage $P^{\min}_a$, and maximum adjustment percentage $P^{\max}_a$. Specifically, OLS first truncates pixels with a large deviation from the centre value through the $P^{\min}$ and $P^{\max}$, and then obtains the truncated minimum pixel value $I_t^{\min}$ and maximum pixel value $I_t^{\max}$ image. Finally, we refine the chosen minimum/maximum pixel values by minimum/maximum adjustment percentages, i.e.,
    \begin{equation}\label{eqols}
        \begin{cases}
            I_{tp}^{\min} = I_{t}^{\min} - P_a^{\min} * (I_{t}^{\max} - I_{t}^{\min}) \\
            I_{tp}^{\max} = I_{t}^{\max} + P_a^{\max} * (I_{t}^{\max} - I_{t}^{\min}) 
        \end{cases},
    \end{equation}
    where $I_{tp}^{\min}$, $I_{tp}^{\max}$ are the optimized pixel minimum and maximum values. In this work, $P^{\min} = 0.01$, $P^{\max} = 0.99$, $P^{\min}_a$ and $P^{\max}_a$ are assigned different values for different degraded scenes, so as to better control the color and texture features of the restored image. Therefore, the OLS-guided image $I_{ols}$ can be given as
    \begin{equation}
        I_{\text{ols}}(x)=\frac{I(x)-I_{tp}^{\min}} {I_{tp}^{\max}-I_{tp}^{\min}}=\textbf{OLS}(I(x)).
    \end{equation}
    Similar to Eq. \ref{eq:dem}, to reduce the loss of texture detail features caused by OLS, we still suggest two SRBs for feature learning and compensation. Therefore, the proposed color restoration module (CRM, $\textbf{CRM}(\cdot)$) can be given as
    \begin{equation}\label{eq:crm}
        \textbf{CRM}(I(x))=\textbf{SRB}(\textbf{OLS}(\mathcal{C}(I(x)))).
    \end{equation}
    The proposed CRM greatly improves the sensitivity of the network to degraded image color and contrast.
\subsection{MRF Extraction Module}
    In degraded images, important features such as edges and colors are often damaged or obscured. To address this issue, we extract features of multi-receptive fields (MRF, $\textbf{MRF}(\cdot)$) with 4 parallel atrous convolution blocks to capture enhanced contextual information, which improves the performance of repairing degraded images. For the input $x_{in}^{m}$, the relevant operations involved in this process can be described as
    \begin{equation}
        \textbf{MRF}(x_{in}^{m}) = Cat[\mathcal{C}_3(x_{in}^{m}),\mathcal{C}_6(x_{in}^{m}),\mathcal{C}_9(x_{in}^{m}),\mathcal{C}_{12}(x_{in}^{m})],
    \end{equation}
    where $Cat[\cdot]$ represents feature concatenation, $\mathcal{C}_3$, $\mathcal{C}_6$, $\mathcal{C}_9$, and $\mathcal{C}_{12}$ are atrous convolution operations with rates of $3$, $6$, $9$ and $12$, respectively. Therefore, the proposed MRF extraction module (MEM $\textbf{MEM}(\cdot)$) can be given as
    \begin{equation}\label{eq:mem}
        \textbf{MEM}(I(x))=\textbf{SRB}(\textbf{MRF}(\mathcal{C}(I(x)))).
    \end{equation}
    The MEM can also assist in initial feature extraction so as to weaken the loss and destruction of initial local features caused by GC and OLS.
\subsection{Encoder-Decoder-based Fusion Module}
    The powerful feature extraction and mapping capabilities of Encoder-Decoder networks have been widely verified \citep{liu2023aioenet}. Therefore, we take SRB as the basic learning block to build an Encoder-Decoder fusion-based module (EDFM) and optimize global feature learning through skip connections. As shown in Fig. \ref{Figure_Flowchart}, EDFM will perform refined fusion learning on three types of features (i.e., EDM-, CRM-, and MEM-mapped). To fully extract the three types of features, we suggest reconstructing and fusing the features through feature addition and feature concatenation. The channel numbers of the three scales of EDFM are $16$, $32$, and $64$, respectively. Max pooling and bilinear interpolation are used for downsampling and upsampling. Skip connections can improve the phenomenon of vanishing gradients and network degradation. EDFM can strengthen the extraction and learning of multi-scene degenerate features (e.g., colors and texture details) while weakening the initial local feature loss brought about by GC and OLS.
\subsection{Loss Function}
    To achieve a good  balance between visual quality and quantitative scores, we use the linear combination of the $\ell_1$-norm loss $\mathcal{L}_{\ell_1}$, contrastive regularization loss $\mathcal{L}_{cr}$, and color loss $\mathcal{L}_{\text {color}}$ as the total loss $\mathcal{L}_{\text {total}}$, which can be expressed as
    \begin{equation}
        \mathcal{L}_{\text {total}}= \lambda_1\mathcal{L}_{\ell_1} + \lambda_2\mathcal{L}_{\text {color}} + \lambda_3\mathcal{L}_{\text {rc}},
        \label{eq:totalloss}
    \end{equation}	
    where $\lambda_1$, $\lambda_2$, and $\lambda_3$ denote the weight value for each loss term, respectively. Extensive experimental results show that $\lambda_1=0.8$, $\lambda_2=0.1$, and $\lambda_3=0.1$ will have the best quantitative and qualitative performance.
\subsubsection{$\ell_1$-norm Loss}
    The loss $\mathcal{L}_{\ell_1}$ can promote the network to learn sparse solutions and exhibit good gradient properties. To verify the effect improvement of our AoSRNet for multi-scene recovery, loss $\mathcal{L}_{\ell_1}$ aims to minimize the $\ell_1$-norm between ground truth $I_g$ and the recovered image $I_r$ through AoSRNet, i.e.,
    \begin{equation}
        \mathcal{L}_{\ell_1} = \min \|I_r- I_g\|_1.
    \end{equation}
\subsubsection{Color Loss}
    The color loss is beneficial to weaken the color distortion caused by the deep network, which can guide AoSRNet to generate images with natural colors and normal contrast \citep{lim2020dslr}. Therefore, to balance the sensitivity to color during multi-scene recovery learning and inference, we suggest introducing a color loss $ \mathcal{L}_{\text {color}}$, which can be given as
    \begin{equation}
        \mathcal{L}_{\text {color}}=1-\frac{\left\langle I_r, I_g\right\rangle}{\left\|I_r\right\|_2 \times\left\|I_g\right\|_2},
    \end{equation}
    where $\left\langle , \right\rangle$ indicates the inner product. By considering the direction of the color vector in the restoration process, the closer the restored image $I_r$ is to the color of the ground true $I_g$, the closer the $\mathcal{L}_{\text {color}}$ is to zero.
\subsubsection{Contrastive Regularization Loss}
    Contrastive regularization (CR) pulls out "positive" pairs in some metric space and separates representations between negative pairs, leading the network to generate better restored images \citep{yin2022degradation}. The mathematical representation of the contrastive loss is
    \begin{equation}
        \mathcal{L}_{cr}= \sum_{i=1}^n \omega_i \cdot \frac{\left\|\psi_i\left(I_r\right)-\psi_i(I_g)\right\|_1}{\left\|\psi_i(I)-\psi_i(I_g)\right\|_1},
    \end{equation}
    where $\psi_i(\cdot),. i=1,2, \cdots n$, refer to extracting the $i$-th hidden features from the VGG-19 network pre-trained. $I_r$, $I$, and $I_g$ are the images restored by AoSRNet, the degraded image, and the real value. $\omega_i$ are weight coefficients, and we set $\omega_1=\frac{1}{32}, \omega_2=\frac{1}{16},\omega_3=\frac{1}{8},\omega_4=\frac{1}{4}$, and $\omega_5=1$.
\section{Experiments and Discussion}\label{sec:experiments}
    In this section, we perform extensive experiments to demonstrate the remarkable low-visibility enhancement performance of AoSRNet. Firstly, we provide an overview of the train/test datasets and explain implementation details. To showcase the superiority of our method, we quantitatively and qualitatively compare it with several state-of-the-art methods using synthetic and real-world three types of low-visibility images. Furthermore, we conduct ablation studies to analyze the parameter value of OLS and the significance of each module.
\subsection{Dataset and Implementation Details}
    \setlength{\tabcolsep}{4.50pt}
    \begin{table}[t]
        \centering
        \scriptsize
        \caption{The details of datasets used in our experiment.}
        \begin{tabular}{l|cc|c|ccc}
            \hline
            Datasets                            & Train & Test  & Depth& Haze         & Sand       & Low-light \\ \hline \hline
            RESIDE-OTS \citep{li2019benchmarking}   & 1200  & 200 & \CheckmarkBold  & \CheckmarkBold & \CheckmarkBold  &\CheckmarkBold\\
            RESIDE-SOTS \citep{li2019benchmarking}  & 0  & 200  & &  & \CheckmarkBold  &\\
            SMD \citep{prasad2017video}             & 600  & 200  &     & \CheckmarkBold                              &    &  \CheckmarkBold    \\\hline
        \end{tabular}
        \label{Table_datasets}
    \end{table}
    \setlength{\tabcolsep}{5.00pt}
    \begin{table}[t]
        \centering
        \scriptsize
        \caption{Methods for comparison with AoSRNet.}
        \begin{tabular}{l|c|c|ccc}
            \hline
            Methods	&       Publication  & Learning       & Haze                          & Sand                                             & Low Light                      \\ \hline\hline
            Fusion \citep{fu2014fusion} & MMSP (2014)  & &                        & \CheckmarkBold &                                               \\
            Retinex \citep{fu2014retinex} & ICIP (2014) &                           &  & \CheckmarkBold &                        \\ 
            CBF \citep{ancuti2017color}    & TIP (2017) &                           &  & \CheckmarkBold &                        \\ 
            SDD \citep{hao2020low}    & TMM (2020) &  &  \CheckmarkBold                         &                        & \CheckmarkBold \\ 
            CCDID \citep{dhara2020color}  & TCSVT (2021)&   & \CheckmarkBold & \CheckmarkBold &                                                \\ 
            Kind+ \citep{zhang2021beyond}  & IJCV (2021) & \CheckmarkBold  &                                                &                        & \CheckmarkBold \\ 
            Ako \citep{bartani2022adaptive} & MTA (2022) &  &                        &    \CheckmarkBold                    &                          \\ 
            ACDC \citep{zhang2022underwater}   & JOE (2022) &    & \CheckmarkBold & & \CheckmarkBold \\ 
            CEEF \citep{liu2021joint}  & TMM (2022) &    & \CheckmarkBold &                                                & \CheckmarkBold \\ 
            ROP+ \citep{liu2022rank}   & TPAMI (2023) &  & \CheckmarkBold & \CheckmarkBold  & \CheckmarkBold \\ 
            PCDE \citep{zhang2023underwater} & SPL (2023) &   & \CheckmarkBold &  & \CheckmarkBold                       \\ 
            SMNet \citep{lin2023smnet}  & TMM (2023) &  \CheckmarkBold  &                        &                        &                         \CheckmarkBold \\ 
            TOENet \citep{gao2023let} & TIM (2023) & \CheckmarkBold    & \CheckmarkBold & \CheckmarkBold &                                                \\ 		\hline
            AoSRNet    & ---           & \CheckmarkBold & \CheckmarkBold & \CheckmarkBold & \CheckmarkBold \\
            \hline
        \end{tabular}
        \label{C-Methods}
    \end{table}
\subsubsection{Train and Test Datasets}
    The application scenes discussed in this work primarily focus on imaging terrestrial and oceanic environments. As shown in Table \ref{Table_datasets}, the training dataset comprises RESIDED-OTS (which incorporates depth information) \citep{li2019benchmarking} for land scenes, the Singapore maritime dataset (SMD) \citep{prasad2017video} for water scenes. To generate more realistic degraded images, including haze and sand dust, we extract atmospheric light values from real-world conditions. These values, in conjunction with an atmospheric scattering model, are used to generate the degraded images. Specifically, we carefully select atmospheric light values for haze and sand dust (named \textbf{AoSRNet-A}), and release them for future scene restoration work. The synthesis of low-visibility image is achieved through the application of the Gamma transformation and Retinex theory. The synthesis of degraded images that closely resemble real-world conditions can address the limitation of insufficient paired training datasets, hence enhancing the restoration performance of the network. When evaluating the inference capabilities of AoSRNet, we employ a range of synthetic and real-world low-visibility images. Specifically, we consider three types of datasets: RESIDED-OTS, RESIDED-SOTS, and SMD. RESIDED-OTS encompasses hazy, sandy, and low-light images, while RESIDED-SOTS focuses on sand dust. SMD covers hazy and low-light scenes.
\subsubsection{Competitive Methods}
    To evaluate the effectiveness of the proposed method, we conduct a comparative analysis between AoSRNet and various state-of-the-art methods, as shown in Table \ref{C-Methods}. The methods being compared encompass both traditional and learning-based, with the majority of them demonstrating the ability to recover low-visibility images in at least two different types of scenes. To ensure the integrity and objectivity of the experiments, all the compared methods are obtained from the source code released by the author.
\subsubsection{Evaluation Metrics}
    To conduct a quantitative evaluation of the recovery performance of the different methods, we have selected a set of evaluation metrics. These metrics include reference evaluation metrics (i.e., peak signal-to-noise ratio (PSNR) and structural similarity (SSIM), and the no-reference evaluation metric (i.e., natural image quality evaluator (NIQE)), which serve as quantitative measures for evaluating the effectiveness of the enhancement methods under consideration. A higher value of PSNR and SSIM indicates superior performance in image recovery. Conversely, the NIQE has an inverse relationship, where a lower value signifies greater recovery performance. It should be noted that all evaluation metric values in this paper are derived based on the RGB channels of the images.
\subsubsection{Experiment Platform}
    The AoSRNet is trained for 100 epochs with 1800 images. The adaptive moment estimation (ADAM) optimizer is responsible for updating the network parameters. The initial learning rate for the AoSRNet is set at $1\times10^{-3}$ and then reduced by a factor of 10 at the 30th, 60th, and 90th epochs. The AoSRNet was trained and evaluated within the Python 3.7 environment using the PyTorch package with 2 Xeon Gold 37.5M Cache, 2.50 GHz @2.30GHz Processors and 4 Nvidia GeForce RTX 4090 GPUs.
\subsection{Quantitative Analysis and Comparison}
\subsubsection{Haze}
    Table \ref{tablehaze} displays the objective evaluation metrics for various image dehazing methods on the RESIDE-OTS dataset for land scenes and the SMD for sea scenes. SDD attempts to improve image quality by transforming hazy images into pseudo-low-light images. However, the resulting restored images fail to meet the desired imaging requirements due to noticeable disparities between the two images. ROP+ refines the transmittance estimation, leading to improved adaptability to various hazy situations and a higher evaluation index value. 
   \setlength{\tabcolsep}{4.5pt}
    \begin{table*}[t]
        \centering
        \scriptsize
        \caption{PSNR, SSIM, and NIQE results of various dehazing methods on RESIDE-OTS \citep{li2019benchmarking} and SMD \citep{prasad2017video}.}
        \begin{tabular}{l|ccc|ccc}
            \hline
             & {PSNR $\uparrow$}             & {SSIM $\uparrow$}                       & NIQE $\downarrow$           & {PSNR $\uparrow$}             & {SSIM $\uparrow$}                       & NIQE $\downarrow$           \\ \hline\hline
            & \multicolumn{3}{c|}{RESIDE-OTS \citep{li2019benchmarking}}                                                                                                       & \multicolumn{3}{c}{SMD \citep{prasad2017video}}                                                                                                              \\ \hline           
            SDD \citep{hao2020low}     & {14.270$\pm$2.667} & {0.797$\pm$0.115} & 4.178$\pm$1.408 & {13.709$\pm$2.444} & {0.820$\pm$0.092}  & 6.647$\pm$0.959 \\ 
            CCDID \citep{dhara2020color}   & {16.168$\pm$3.424} & {0.827$\pm$0.063}  & 3.503$\pm$0.996 & {17.381$\pm$3.178} & {0.907$\pm$0.042} & 5.948$\pm$1.410 \\ 
            ACDC \citep{zhang2022underwater}    & {17.345$\pm$2.402} & {0.743$\pm$0.062} & 3.419$\pm$0.899 & {18.196$\pm$1.925} & {0.745$\pm$0.057}  & 5.762$\pm$1.533 \\ 
            CEEF \citep{liu2021joint}    & {13.837$\pm$2.586} & {0.789$\pm$0.067}  & 3.458$\pm$0.892 & {12.723$\pm$2.637} & {0.854$\pm$0.062} & 5.581$\pm$0.981 \\ 
            ROP+ \citep{liu2022rank}    & {19.118$\pm$3.604} & {0.865$\pm$0.059}  & 3.518$\pm$1.040 & {18.402$\pm$2.945} & {0.883$\pm$0.048}  & 6.289$\pm$1.550 \\ 
            PCDE \citep{zhang2023underwater}   & {16.446$\pm$2.877} & {0.777$\pm$0.062}  & 3.684$\pm$1.275 & {14.276$\pm$2.288} & {0.755$\pm$0.098}  & 6.960$\pm$4.777 \\ 
            TOENet \citep{gao2023let}  & {22.729$\pm$4.185} & {0.903$\pm$0.052} & {0.983$\pm$0.012}  & {18.619$\pm$3.289} & {0.891$\pm$0.077}  & 5.522$\pm$1.092 \\ \hline
            AoSRNet & {22.939$\pm$4.575} & {0.910$\pm$0.054}  & 3.328$\pm$1.025 & {20.030$\pm$3.641} & {0.962$\pm$0.033}  & 5.826$\pm$1.145 \\ \hline
        \end{tabular}\label{tablehaze}
    \end{table*}
    \setlength{\tabcolsep}{4.5pt}
    \begin{table*}[t]
        \centering
        \scriptsize
        \caption{PSNR, SSIM, and NIQE results of various sandy enhancement methods on RESIDE-OTS \citep{li2019benchmarking} and RESIDE-SOTS \citep{li2019benchmarking}}
        \begin{tabular}{l|ccc|ccc}
            \hline
             & {PSNR $\uparrow$}             & {SSIM $\uparrow$}                       & NIQE $\downarrow$           & {PSNR $\uparrow$}             & {SSIM $\uparrow$}                       & NIQE $\downarrow$           \\ \hline\hline            
            & \multicolumn{3}{c|}{RESIDE-OTS\citep{li2019benchmarking}  }                                                                                                       & \multicolumn{3}{c}{RESIDE-SOTS \citep{li2019benchmarking} }    \\ \hline
            Fusion \citep{fu2014fusion}  & {18.518$\pm$3.248} & {0.781$\pm$0.103}  & 4.557$\pm$1.870 & {19.719$\pm$4.844} & {0.800$\pm$0.151}  & 3.609$\pm$1.077 \\ 
            Retinex \citep{fu2014retinex} & {16.965$\pm$2.742} & {0.752$\pm$0.081}  & 3.969$\pm$0.793 & {18.208$\pm$2.594} & {0.807$\pm$0.059}  & 3.500$\pm$0.611 \\ 
            CBF \citep{ancuti2017color}     & {13.419$\pm$2.491} & {0.749$\pm$0.093}  & 3.712$\pm$0.849 & {13.446$\pm$2.362} & {0.773$\pm$0.098}  & 3.342$\pm$0.747 \\ 
            CCDID \citep{dhara2020color}   & {12.028$\pm$3.283} & {0.735$\pm$0.096}  & 3.746$\pm$1.093 & {12.874$\pm$3.439} & {0.788$\pm$0.085}  & 3.262$\pm$0.985 \\ 
            Ako \citep{bartani2022adaptive}     & {15.017$\pm$2.812} & {0.757$\pm$0.072}  & 3.323$\pm$0.819 & {17.283$\pm$3.378} & {0.804$\pm$0.086}  & 2.886$\pm$0.692 \\
            ROP+ \citep{liu2022rank}     & {15.284$\pm$3.629} & {0.790$\pm$0.087}  & 3.689$\pm$1.116 & {18.717$\pm$5.976} & {0.849$\pm$0.101}  & 3.071$\pm$0.764 \\
            TOENet \citep{gao2023let}  & {18.839$\pm$4.421} & {0.851$\pm$0.080}  & 3.546$\pm$0.980 & {21.363$\pm$4.843} & {0.901$\pm$0.089}  & 3.132$\pm$0.688 \\ \hline
            AoSRNet & {19.341$\pm$3.441} & {0.858$\pm$0.073}  & 3.500$\pm$0.954 & {22.134$\pm$4.922} & {0.909$\pm$0.065} & 3.115$\pm$0.738 \\ \hline
        \end{tabular}\label{tablesand}
    \end{table*}
    \setlength{\tabcolsep}{4.5pt}
    \begin{table*}[!ht]
        \centering
        \scriptsize
        \caption{PSNR, SSIM, and NIQE results of various low-light enhancement methods on RESIDE-OTS \citep{li2019benchmarking} and SMD \citep{prasad2017video}.}
        \begin{tabular}{l|ccc|ccc}
            \hline
             & {PSNR $\uparrow$}             & {SSIM $\uparrow$}                       & NIQE $\downarrow$           & {PSNR $\uparrow$}             & {SSIM $\uparrow$}                       & NIQE $\downarrow$           \\ \hline\hline
            & \multicolumn{3}{c|}{RESIDE-OTS \citep{li2019benchmarking}}                                                                                                       & \multicolumn{3}{c}{SMD \citep{prasad2017video}}                                                                                                              \\ \hline
            SDD \citep{hao2020low}      & {16.304$\pm$3.238} & {0.784$\pm$0.131} & 4.670$\pm$1.376 & {15.182$\pm$3.407} & {0.853$\pm$0.094} & 6.505$\pm$0.926 \\ 
            KinD+ \citep{zhang2021beyond} & {15.792$\pm$2.263} & {0.643$\pm$0.120} & 4.019$\pm$0.995 & {18.286$\pm$3.076} & {0.899$\pm$0.063} & 5.887$\pm$0.919 \\ 
            ACDC \citep{zhang2022underwater}     & {18.601$\pm$2.868} & {0.738$\pm$0.078} & 3.458$\pm$0.950 & {16.892$\pm$2.729} & {0.700$\pm$0.071} & 4.755$\pm$0.804 \\ 
            CEEF \citep{liu2021joint}     & {10.216$\pm$2.093} & {0.569$\pm$0.147} & 3.832$\pm$1.003 & {7.611$\pm$1.123}  & {0.534$\pm$0.076} & 5.326$\pm$0.790 \\ 
            ROP+ \citep{liu2022rank}      & {19.284$\pm$4.400} & {0.812$\pm$0.106} & 3.874$\pm$1.057 & {16.468$\pm$3.636} & {0.844$\pm$0.072} & 5.464$\pm$1.023 \\ 
            PCDE \citep{zhang2023underwater}    & {13.955$\pm$2.408} & {0.511$\pm$0.133} & 4.374$\pm$1.285 & {11.193$\pm$1.976} & {0.606$\pm$0.129} & 4.029$\pm$0.546 \\
            SMNet \citep{lin2023smnet}    & {16.459$\pm$2.762} & {0.800$\pm$0.107} & 4.436$\pm$1.446 & {14.621$\pm$3.730} & {0.848$\pm$0.087} & 5.983$\pm$0.785 \\ \hline
            AoSRNet  & {26.691$\pm$4.436} & {0.924$\pm$0.044} & 3.742$\pm$1.110 & {25.139$\pm$5.119} & {0.981$\pm$0.013} & 5.762$\pm$0.990 \\ \hline
        \end{tabular}\label{tablelow}
    \end{table*}
    Since there is more land scene data in the training set, the performance of the learning-driven TOENet in the land scene is better than that in the water scene, which reflects that the learning-based method often has a high dependence on the training dataset. In comparison, the utilization of DEM and CRM enabled AoSRNet to evolve beyond exclusive reliance. These modules enable the preprocessing of degraded images using both nonlinear and linear transformations and exhibit robust scene generalization capabilities. Hence, AoSRNet obtained better rankings in different evaluation metrics.
    \begin{figure*}[tb]
        \centering
        \setlength{\abovecaptionskip}{0.cm}
        \includegraphics[width=1.00\linewidth]{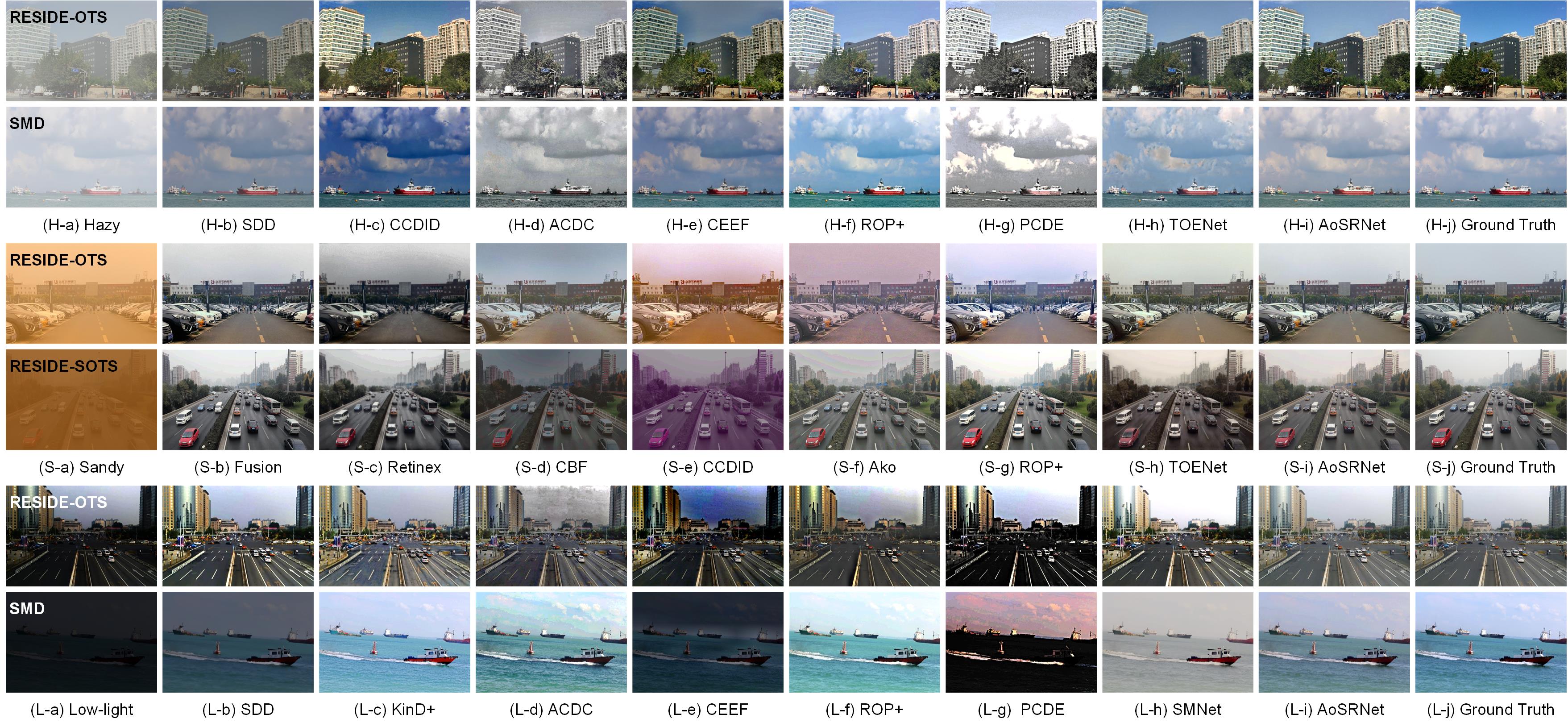}
        \caption{Visual comparisons of scene recovery performance from synthetic three types of low-visibility images.}
        \vspace{-0.3cm}
        \label{Figuresyn}
    \end{figure*}
\subsubsection{Sand Dust}
    Sand dust presents a greater level of complexity compared to haze, primarily due to the presence of inconsistent atmospheric light levels. This characteristic poses challenges for the application of DCP-based methods in sand dust image enhancement. To address this issue, most sand dust image enhancement methods employ GC or LS as either pre-processing or post-processing procedures. As shown in Table \ref{tablesand}, Fusion successfully combines GC and LS while disregarding the atmospheric scattering model, making it effective for generating synthetic sandstorm images. However, in instances where the degradation scene shows a high level of complexity, the performance of Fusion is expected to see a substantial decline. The Retinex can effectively decompose degraded images and shows satisfactory recovery performance. CBF and CCDID lack the ability for dense sandstorm scene recovery, leading to poor evaluation index values. The metric values of Ako and ROP+ show a degree of similarity, but their performance in complex sand and sand dust scenes is deemed inadequate. The learning-based methods demonstrate a comparatively elevated assessment metric, with AoSRNet showing superior stability in comparison to TOENet.
    \begin{figure*}[tb]
        \centering
        \setlength{\abovecaptionskip}{0.cm}
        \includegraphics[width=1.00\linewidth]{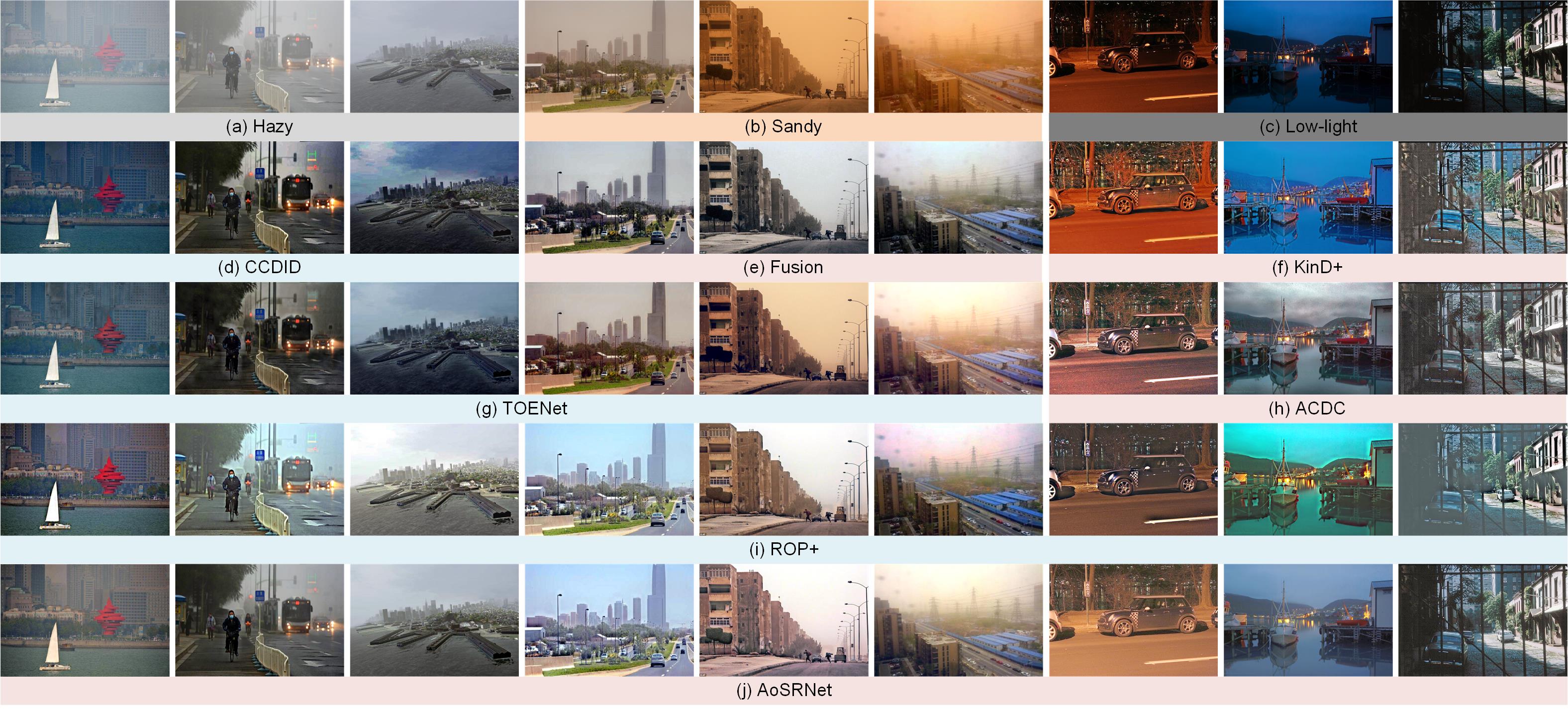}
        \caption{Visual comparisons of scene recovery performance from real-world three types of low-visibility images.}
        \vspace{-0.3cm}
        \label{Figurereal}
    \end{figure*}
\subsubsection{Low Light}
    The restoration ability of AoSRNet is demonstrated in Table \ref{tablelow}. It can be seen from the evaluation indicators that compared to other competing methods, AoSRNet exhibits significant advantages in restoring low-visibility images that are synthesized using Retinex theory and Gamma transformation. Traditional low-light enhancement methods are generally more robust across different scenes. However, complex low-light degraded images often deviate from the original imaging model, leading to poor enhancement performance of traditional methods, especially in scenarios with uneven illumination or multiple light sources. It is also important to note that learning-based methods may have limitations in terms of over-reliance on the training dataset, which can result in insufficient scene generalization ability. In comparison, In comparison, AoSRNet can combine the advantages of traditional and learning methods and weaken the shortcomings, achieving satisfactory performance.
\subsection{Visual Analysis and Comparison}
\subsubsection{Visual Comparison with Reference}
    We conducted a visual analysis of degraded images with references to assess the generalization ability of AoSRNet in different scenes. To evaluate its performance, as shown in Fig. \ref{Figuresyn}, we selected classic low-visibility images of land and ocean under various degradation scenes. The RESIDE-OTS, which includes depth information, was observed to produce low-visibility images that closely resembled real degraded images. However, traditional methods struggle to obtain depth information on the degraded image, resulting in either over-enhancement of local areas and loss of texture details or under-enhancement, allowing degradation factors to persist. We also conducted fair tests on RESIDE-SOTS and SMD, where depth information was not considered. Most methods still faced challenges in accurately extracting valuable information from these degraded images. The complexity of the imaging environment often deviates from the original degradation imaging model, making it difficult for traditional methods to achieve satisfactory visual restoration performance in different degradation scenes. Additionally, learning-based methods rely heavily on training data and lack generalization ability, making achieving good visual restoration results in unpredictable imaging degradation environments challenging. However, AoSRNet combines the advantages of network models and physical prior information, enabling it to achieve satisfactory visual performance in various degraded environments.
    \begin{figure}[tb]
        \centering
        \setlength{\abovecaptionskip}{0.cm}
        \includegraphics[width=0.85\linewidth]{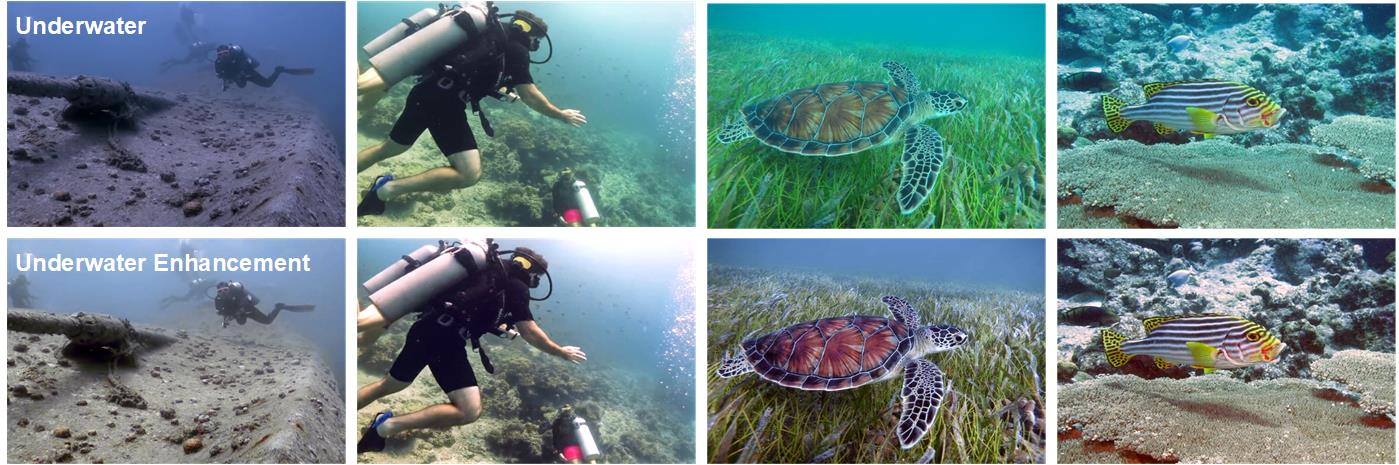}
        \caption{Visual results of our method for underwater image enhancement.}
        \vspace{-0.3cm}
        \label{Figureunder}
    \end{figure}
\subsubsection{Visual Comparison without Reference}
    This subsection focuses on restoring real-world degraded images without any references. We selected the top-performing methods for a comparative analysis. As shown in Fig. \ref{Figurereal}, it is worth noting that real-world imaging degradation is significantly more intricate and challenging to estimate accurately. Traditional methods often struggle with issues such as over-enhancement or under-enhancement. In contrast, learning methods risk overfitting the data, resulting in unsatisfactory visual performance of the restored image in both global and local areas. However, compared to competing methods, AoSRNet demonstrates strong scene generalization ability and restoration robustness, delivering the best visual performance.
\subsubsection{Generalization Performance of AoSRNet}
    The proposed all-in-one scene recovery network generalizes well to underwater image enhancement without any parameter fine-tuning. As shown in the bottom row of Fig. \ref{Figureunder}, the images enhanced by AosRNet are visually pleasing, meanwhile, the contrast and details are well enhanced. Moreover, the details and visibility are well enhanced. These enhanced results are yielded by our method without parameters fine-tuning, which demonstrates the good generalization performance of our method.
    \begin{figure}[tb]
        \centering
        \setlength{\abovecaptionskip}{0.cm}
        \includegraphics[width=0.85\linewidth]{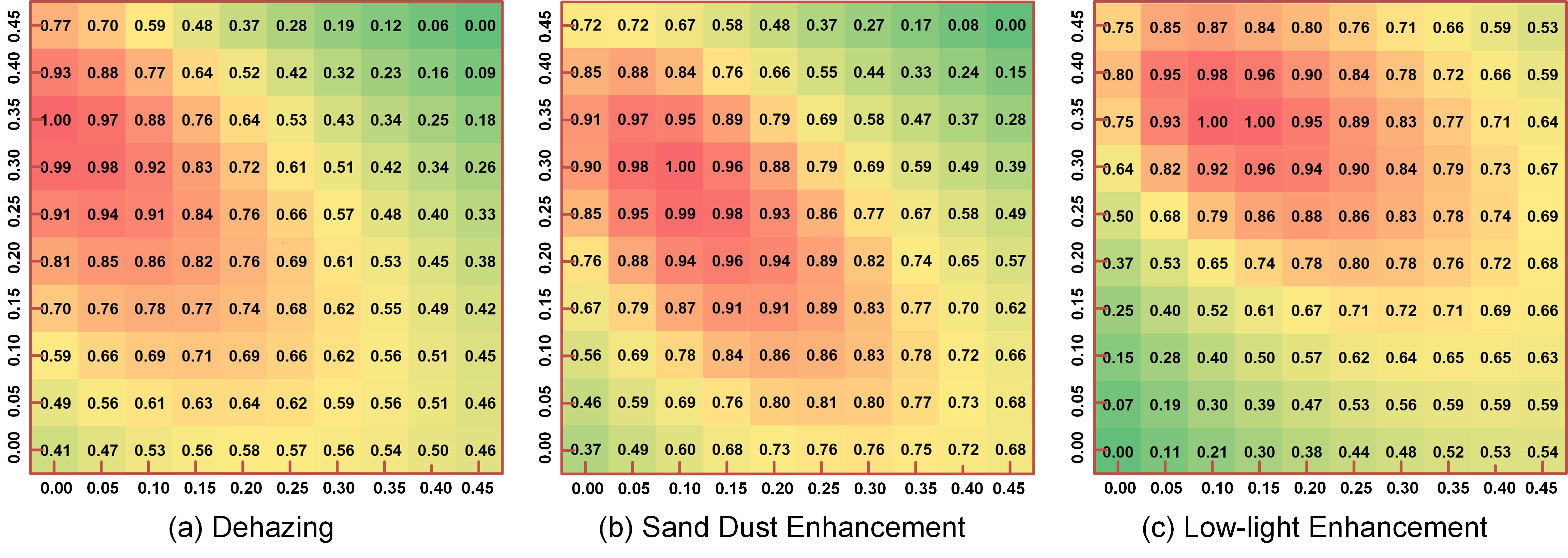}
        \caption{The impact of the values of $P_a^{\min}$ and $P_a^{\max}$ in the Eq. \ref{eqols} on the performance of AoSRNet, where $1$ means the best performance, and $0$ means the worst performance.}
        \vspace{-0.3cm}
        \label{Figureols}
    \end{figure}
\subsection{Ablation Study}\label{ss:as}
\subsubsection{Parametric Analysis of OLS}
    Different degradation factors have different parameter requirements for the suggested OLS, and a better restoration effect can be achieved by fine-tuning the parameters of the OLS. To this end, this part will adjust the OLS suitable for the three degradation scenes (i.e., $P^{\min}$, $P^{\max}$, $P^{\min}_a$, and $P^{\max}_a$ in the Eq. \ref{eqols}) to obtain the best parameters. We first determine $P^{\min}$ and $P^{\max}$ as the general $0.01$ and $0.99$, then mainly perform linear transformation on $P^{\min}_a$ and $P^{\max}_a$, and finally make statistics on the objective evaluation indicators of the restored image. As shown in Fig. \ref{Figureols}, the optimal choices for  $P^{\min}_a$ and $P^{\max}_a$ are significantly different for the three degradation scenes. Therefore, to better optimize AoSRNet, we selected three optimal ($P^{\min}_a$ , $P^{\max}_a$) embedding networks and performed joint optimization.
\subsubsection{Effect of DEM and CRM}\label{as:demandcrm}
    We conduct ablation experiments to verify the significance of DEM and CRM for AoSRNet to learn inference. As shown in Table \ref{tabledemcrm}, when AoSRNet lacks the auxiliary learning of DEM and CRM, its performance is the worst among the three restoration tasks. Furthermore, by respectively embedding DEM and CRM into the main network, we observed improvements in the objective evaluation index values of AoSRNet. And when AoSRNet integrates both DEM and CRM, it exhibits the best restoration performance. This highlights the synergistic effect of these components and their combined contribution towards improving the overall performance of AoSRNet. Additional nonlinear and linear transformations reduce the risk of the network being prone to overfitting when propagating learning, and gamma correction and optimized linear stretching also provide AoSRNet with rich and valuable prior feature information. 
    \setlength{\tabcolsep}{6.00 pt}
    \begin{table}[tb]
        \centering
        \scriptsize
        \caption{Ablation study on different modules in AoSRNet.}
        \begin{tabular}{ccc|cc}
            \hline
            MEM & CRB & DEM & PSNR  $\uparrow$             & SSIM $\uparrow$          \\ \hline\hline
            &    & &{20.363$\pm$3.926} & 0.903$\pm$0.071   \\ 
            \CheckmarkBold   &    & & {20.983$\pm$3.409} & 0.908$\pm$0.050  \\ 
            \CheckmarkBold & \CheckmarkBold   & & {20.993$\pm$3.497} & 0.913$\pm$0.051  \\ 
            \CheckmarkBold &  &  \CheckmarkBold & {21.202$\pm$3.038} & 0.912$\pm$0.047 \\
        & \CheckmarkBold & \CheckmarkBold  & {21.306$\pm$3.773} & 0.915$\pm$0.063 \\\hline
            \CheckmarkBold   & \CheckmarkBold & \CheckmarkBold & {21.475$\pm$3.732} & 0.918$\pm$0.058 \\ \hline
        \end{tabular}\label{tabledemcrm}
    \end{table}
\subsubsection{Effect of Loss Function}\label{as:lossfunction}
    A suitable loss function is an important part of ensuring that a deep network model works in the expected way. To this end, this subsection conducts ablation learning on the proposed three loss functions to verify the important role each loss function plays in the learning process of AoSRNet. As shown in Table \ref{tableloss}, on the basis of $\ell_1$-norm loss $\mathcal{L}_{\ell_1}$ as the basic loss function, color loss $\mathcal{L}_{\text {color}}$ and contrastive regularization loss $\mathcal{L}_{\text {rc}}$ can improve the restoration performance respectively, and when the three loss functions are available at the same time, the convergence speed and restoration performance of the network reach the best excellent condition.
    \setlength{\tabcolsep}{6.00 pt}
    \begin{table}[tb]
        \centering
        \scriptsize
        \caption{Ablation study on different loss function in AoSRNet.}
        \begin{tabular}{ccc|cc}
            \hline
            $\mathcal{L}_{\ell_1}$ & $\mathcal{L}_{\text {color}}$ & $\mathcal{L}_{\text {rc}}$ & PSNR  $\uparrow$             & SSIM $\uparrow$          \\ \hline\hline
            \CheckmarkBold &    & &{20.093$\pm$4.114} & 0.894$\pm$0.066   \\ 
            \CheckmarkBold   & \CheckmarkBold   & & {20.450$\pm$3.764} & 0.903$\pm$0.061  \\ 
            \CheckmarkBold &    & \CheckmarkBold& {20.655$\pm$3.672} & 0.908$\pm$0.055  \\ 
            \CheckmarkBold   & \CheckmarkBold & \CheckmarkBold & {21.475$\pm$3.732} & 0.918$\pm$0.058 \\ \hline
        \end{tabular}\label{tableloss}
    \end{table}
\subsection{Limitation}
    In this work, the parameters of OLS and GC (i.e., $\gamma$, $P^{\min}_a$, and $P^{\max}_a$) are preset. However, different degradation scenarios will only have an optimal set of OLS and GC parameter combinations. Although AoSRNet presets four sets of parameters respectively, and performs learning and reasoning through deep networks. However, inappropriate OLS and GC in specific imaging scenarios may still have a negative impact on the final restoration results. Therefore, how to further optimize OLS and GC to more robustly adapt to complex imaging environments is what AoSRNet needs to continue to study in depth.
\section{Conclusion}\label{sec:conclusions}
    In this work, we propose a general all-in-one scene recovery network via multi-knowledge integration (termed AoSRNet) to improve the visibility of imaging devices in three common low-visibility imaging scenes (i.e., haze, sand dust, and low light). Specifically, we combine the gamma correction (GC) and optimized linear stretching (OLS) with the standard residual blocks respectively, thus proposing the detail enhancement module (DEM) and color restoration module (CRM) to guide the sub-learning networks to restore the degraded image. In addition, we suggest a multi-receptive field extraction module (MEM) to attenuate the loss of image texture details caused by GC nonlinear and OLS linear transformations. DEM and CRM will alleviate the overfitting of the deep network so that AoSRNet can improve the imaging quality of the visual sensor more robustly and efficiently in different degradation scenes. Finally, we refine and fuse the coarse features generated by DEM, CRM, and MEM through the Encoder-Decoder-based fusion module to generate the final restored image. Comprehensive experimental results demonstrate that our method is efficient and stable compared with other state-of-the-art methods for image restoration tasks in vision-driven intelligent devices and systems.

\bibliographystyle{elsarticle-num}
\footnotesize
\bibliography{KBS}

\begin{thebibliography}{10}
\expandafter\ifx\csname url\endcsname\relax
  \def\url#1{\texttt{#1}}\fi
\expandafter\ifx\csname urlprefix\endcsname\relax\def\urlprefix{URL }\fi
\expandafter\ifx\csname href\endcsname\relax
  \def\href#1#2{#2} \def\path#1{#1}\fi

\bibitem{liu2022rank}
J.~Liu, R.~W. Liu, J.~Sun, T.~Zeng, Rank-one prior: Real-time scene recovery, IEEE Trans. Pattern Anal. Mach. Intell. 45~(7) (2022) 8845--8860.

\bibitem{chen2023mask}
S.~Chen, T.~Shu, H.~Zhao, Y.~Y. Tang, Mask-cnn-transformer for real-time multi-label weather recognition, Knowl-Based Syst. 278 (2023) 110881.

\bibitem{chen2023edge}
C.~Chen, C.~Wang, B.~Liu, C.~He, L.~Cong, S.~Wan, Edge intelligence empowered vehicle detection and image segmentation for autonomous vehicles, IEEE Trans. Intell. Transp. Syst. (Jan. 2023).

\bibitem{he2010single}
K.~He, J.~Sun, X.~Tang, Single image haze removal using dark channel prior, IEEE Trans. Pattern Anal. Mach. Intell. 33~(12) (2010) 2341--2353.

\bibitem{land1977retinex}
E.~H. Land, The retinex theory of color vision, Sci. Am. 237~(6) (1977) 108--129.

\bibitem{kim2019fast}
S.~E. Kim, T.~H. Park, I.~K. Eom, Fast single image dehazing using saturation based transmission map estimation, IEEE Trans. Image Process. 29 (2019) 1985--1998.

\bibitem{jiang2013night}
X.~Jiang, H.~Yao, S.~Zhang, X.~Lu, W.~Zeng, Night video enhancement using improved dark channel prior, in: Proc. IEEE ICIP, 2013, pp. 553--557.

\bibitem{fu2014retinex}
X.~Fu, P.~Zhuang, Y.~Huang, Y.~Liao, X.-P. Zhang, X.~Ding, A retinex-based enhancing approach for single underwater image, in: Proc. IEEE ICIP, 2014, pp. 4572--4576.

\bibitem{kandhway2023adaptive}
P.~Kandhway, An adaptive low-light image enhancement using canonical correlation analysis, IEEE Trans. Ind. Inf. (Jan. 2023).

\bibitem{zhu2020novel}
Z.~Zhu, H.~Wei, G.~Hu, Y.~Li, G.~Qi, N.~Mazur, A novel fast single image dehazing algorithm based on artificial multiexposure image fusion, IEEE Trans. Instrum. Meas. 70 (2020) 1--23.

\bibitem{zhang2022underwater}
W.~Zhang, Y.~Wang, C.~Li, Underwater image enhancement by attenuated color channel correction and detail preserved contrast enhancement, IEEE J. Oceanic Eng. 47~(3) (2022) 718--735.

\bibitem{liu2021joint}
X.~Liu, H.~Li, C.~Zhu, Joint contrast enhancement and exposure fusion for real-world image dehazing, IEEE Trans. on Multimedia 24 (2021) 3934--3946.

\bibitem{jia2021effective}
T.~Jia, J.~Li, L.~Zhuo, G.~Li, Effective meta-attention dehazing networks for vision-based outdoor industrial systems, IEEE Trans. Ind. Inf. 18~(3) (2021) 1511--1520.

\bibitem{li2022single}
Y.~Li, D.~Cheng, D.~Zhang, N.~Wang, X.~Gao, J.~Sun, Single image dehazing with an independent detail-recovery network, Knowl-Based Syst. 254 (2022) 109579.

\bibitem{ma2021learning}
L.~Ma, R.~Liu, J.~Zhang, X.~Fan, Z.~Luo, Learning deep context-sensitive decomposition for low-light image enhancement, IEEE Trans. Neur. Net. Lear. 33~(10) (2021) 5666--5680.

\bibitem{dai2024understanding}
J.~Dai, Q.~Li, H.~Wang, L.~Liu, Understanding images of surveillance devices in the wild, Knowl-Based Syst. 284 (2024) 111226.

\bibitem{wang2023uscformer}
Y.~Wang, J.~Xiong, X.~Yan, M.~Wei, Uscformer: Unified transformer with semantically contrastive learning for image dehazing, IEEE Trans. Intell. Transp. Syst. (Jun. 2023).

\bibitem{zhou2023physical}
H.~Zhou, Z.~Chen, Y.~Liu, Y.~Sheng, W.~Ren, H.~Xiong, Physical-priors-guided dehazeformer, Knowl-Based Syst. 266 (2023) 110410.

\bibitem{croitoru2023diffusion}
F.-A. Croitoru, V.~Hondru, R.~T. Ionescu, M.~Shah, Diffusion models in vision: A survey, IEEE Trans. Pattern Anal. Mach. Intell. (Mar. 2023).

\bibitem{ren2016single}
W.~Ren, S.~Liu, H.~Zhang, J.~Pan, X.~Cao, M.-H. Yang, Single image dehazing via multi-scale convolutional neural networks, in: Proc. ECCV, 2016, pp. 154--169.

\bibitem{zhang2021beyond}
Y.~Zhang, X.~Guo, J.~Ma, W.~Liu, J.~Zhang, Beyond brightening low-light images, Int. J. Comput. Vision 129 (2021) 1013--1037.

\bibitem{zhao2021refinednet}
S.~Zhao, L.~Zhang, Y.~Shen, Y.~Zhou, Refinednet: A weakly supervised refinement framework for single image dehazing, IEEE Trans. Image Process. 30 (2021) 3391--3404.

\bibitem{zhou2022fsad}
Y.~Zhou, Z.~Chen, P.~Li, H.~Song, C.~P. Chen, B.~Sheng, Fsad-net: Feedback spatial attention dehazing network, IEEE Trans. Neur. Net. Lear. (Feb. 2022).

\bibitem{song2023vision}
Y.~Song, Z.~He, H.~Qian, X.~Du, Vision transformers for single image dehazing, IEEE Trans. Image Process. 32 (2023) 1927--1941.

\bibitem{guo2023scanet}
Y.~Guo, Y.~Gao, W.~Liu, Y.~Lu, J.~Qu, S.~He, W.~Ren, Scanet: Self-paced semi-curricular attention network for non-homogeneous image dehazing, in: Proc. IEEE CVPRW, 2023, pp. 1884--1893.

\bibitem{shu2019variational}
Q.~Shu, C.~Wu, Z.~Xiao, R.~W. Liu, Variational regularized transmission refinement for image dehazing, in: Proc. IEEE ICIP, 2019, pp. 2781--2785.

\bibitem{qin2020ffa}
X.~Qin, Z.~Wang, Y.~Bai, X.~Xie, H.~Jia, Ffa-net: Feature fusion attention network for single image dehazing, in: Proc. AAAI, Vol.~34, 2020, pp. 11908--11915.

\bibitem{liu2022deep}
R.~W. Liu, Y.~Guo, Y.~Lu, K.~T. Chui, B.~B. Gupta, Deep network-enabled haze visibility enhancement for visual iot-driven intelligent transportation systems, IEEE Trans. Ind. Inf. 19~(2) (2022) 1581--1591.

\bibitem{li2017aod}
B.~Li, X.~Peng, Z.~Wang, J.~Xu, D.~Feng, Aod-net: All-in-one dehazing network, in: Proc. IEEE ICCV, 2017, pp. 4770--4778.

\bibitem{cheng2020fast}
Y.~Cheng, Z.~Jia, H.~Lai, J.~Yang, N.~K. Kasabov, A fast sand-dust image enhancement algorithm by blue channel compensation and guided image filtering, IEEE Access 8 (2020) 196690--196699.

\bibitem{zhu2018haze}
Y.~Zhu, G.~Tang, X.~Zhang, J.~Jiang, Q.~Tian, Haze removal method for natural restoration of images with sky, Neurocomputing 275 (2018) 499--510.

\bibitem{peng2018generalization}
Y.-T. Peng, K.~Cao, P.~C. Cosman, Generalization of the dark channel prior for single image restoration, IEEE Trans. Image Process. 27~(6) (2018) 2856--2868.

\bibitem{wang2021fast}
B.~Wang, B.~Wei, Z.~Kang, L.~Hu, C.~Li, Fast color balance and multi-path fusion for sandstorm image enhancement, Signal Image Video P. 15 (2021) 637--644.

\bibitem{fu2014fusion}
X.~Fu, Y.~Huang, D.~Zeng, X.-P. Zhang, X.~Ding, A fusion-based enhancing approach for single sandstorm image, in: Proc. IEEE MMSP, 2014, pp. 1--5.

\bibitem{shi2020normalised}
Z.~Shi, Y.~Feng, M.~Zhao, E.~Zhang, L.~He, Normalised gamma transformation-based contrast-limited adaptive histogram equalisation with colour correction for sand--dust image enhancement, IET Image Process. 14~(4) (2020) 747--756.

\bibitem{gao2022color}
G.~Gao, H.~Lai, L.~Wang, Z.~Jia, Color balance and sand-dust image enhancement in lab space, Multimed. Tools Appl. 81~(11) (2022) 15349--15365.

\bibitem{gao2023let}
Y.~Gao, W.~Xu, Y.~Lu, Let you see in haze and sandstorm: Two-in-one low-visibility enhancement network, IEEE Trans. Instrum. Meas. (Aug. 2023).

\bibitem{si2022sand}
Y.~Si, F.~Yang, Z.~Liu, Sand dust image visibility enhancement algorithm via fusion strategy, Sci. Rep. 12~(1) (2022) 13226.

\bibitem{ding2022restoration}
B.~Ding, H.~Chen, L.~Xu, R.~Zhang, Restoration of single sand-dust image based on style transformation and unsupervised adversarial learning, IEEE Access 10 (2022) 90092--90100.

\bibitem{pizer1987adaptive}
S.~M. Pizer, E.~P. Amburn, J.~D. Austin, R.~Cromartie, A.~Geselowitz, T.~Greer, B.~ter Haar~Romeny, J.~B. Zimmerman, K.~Zuiderveld, Adaptive histogram equalization and its variations, Comput. Vis., Graph., Image Process. 39~(3) (1987) 355--368.

\bibitem{reza2004realization}
A.~M. Reza, Realization of the contrast limited adaptive histogram equalization (clahe) for real-time image enhancement, J. Signal Process. Sys. 38 (2004) 35--44.

\bibitem{feng2020low}
X.~Feng, J.~Li, Z.~Hua, Low-light image enhancement algorithm based on an atmospheric physical model, Multimed. Tools Appl. 79~(43-44) (2020) 32973--32997.

\bibitem{wang2013naturalness}
S.~Wang, J.~Zheng, H.-M. Hu, B.~Li, Naturalness preserved enhancement algorithm for non-uniform illumination images, IEEE Trans. Image Process. 22~(9) (2013) 3538--3548.

\bibitem{guo2016lime}
X.~Guo, Y.~Li, H.~Ling, Lime: Low-light image enhancement via illumination map estimation, IEEE Trans. Image Process. 26~(2) (2016) 982--993.

\bibitem{li2018structure}
M.~Li, J.~Liu, W.~Yang, X.~Sun, Z.~Guo, Structure-revealing low-light image enhancement via robust retinex model, IEEE Trans. Image Process. 27~(6) (2018) 2828--2841.

\bibitem{ren2019low}
W.~Ren, S.~Liu, L.~Ma, Q.~Xu, X.~Xu, X.~Cao, J.~Du, M.-H. Yang, Low-light image enhancement via a deep hybrid network, IEEE Trans. Image Process. 28~(9) (2019) 4364--4375.

\bibitem{lu2022mtrbnet}
Y.~Lu, Y.~Guo, R.~W. Liu, W.~Ren, Mtrbnet: Multi-branch topology residual block-based network for low-light enhancement, IEEE Signal Process Lett. 29 (2022) 1127--1131.

\bibitem{xu2023low}
X.~Xu, R.~Wang, J.~Lu, Low-light image enhancement via structure modeling and guidance, in: Proc. IEEE CVPR, 2023, pp. 9893--9903.

\bibitem{wei2018deep}
C.~Wei, W.~Wang, W.~Yang, J.~Liu, Deep retinex decomposition for low-light enhancement, arXiv preprint arXiv:1808.04560 (Aug. 2018).

\bibitem{ancuti2017color}
C.~O. Ancuti, C.~Ancuti, C.~De~Vleeschouwer, P.~Bekaert, Color balance and fusion for underwater image enhancement, IEEE Trans. Image Process. 27~(1) (2017) 379--393.

\bibitem{qu2023deep}
J.~Qu, Y.~Gao, Y.~Lu, W.~Xu, R.~W. Liu, Deep learning-driven surveillance quality enhancement for maritime management promotion under low-visibility weathers, Ocean Coast. Manage. 235 (2023) 106478.

\bibitem{kim2021deep}
G.~Kim, J.~Kwon, Deep illumination-aware dehazing with low-light and detail enhancement, IEEE Trans. Intell. Transp. Syst. 23~(3) (2021) 2494--2508.

\bibitem{liu2023aioenet}
R.~W. Liu, Y.~Lu, Y.~Guo, W.~Ren, F.~Zhu, Y.~Lv, Aioenet: All-in-one low-visibility enhancement to improve visual perception for intelligent marine vehicles under severe weather conditions, IEEE Trans. Intell. Veh. (Dec. 2023).

\bibitem{lim2020dslr}
S.~Lim, W.~Kim, Dslr: Dep stacked laplacian restorer for low-light image enhancement, IEEE Trans. on Multimedia 23 (2020) 4272--4284.

\bibitem{yin2022degradation}
S.~Yin, S.~Hu, Y.~Wang, W.~Wang, C.~Li, Y.-H. Yang, Degradation-aware and color-corrected network for underwater image enhancement, Knowl-Based Syst. 258 (2022) 109997.

\bibitem{li2019benchmarking}
B.~Li, W.~Ren, D.~Fu, D.~Tao, D.~Feng, W.~Zeng, Z.~Wang, Benchmarking single-image dehazing and beyond, IEEE Trans. Image Process. 28~(1) (2019) 492--505.

\bibitem{prasad2017video}
D.~K. Prasad, D.~Rajan, L.~Rachmawati, E.~Rajabally, C.~Quek, Video processing from electro-optical sensors for object detection and tracking in a maritime environment: A survey, IEEE Trans. Intell. Transp. Syst. 18~(8) (2017) 1993--2016.

\bibitem{hao2020low}
S.~Hao, X.~Han, Y.~Guo, X.~Xu, M.~Wang, Low-light image enhancement with semi-decoupled decomposition, IEEE Trans. on Multimedia 22~(12) (2020) 3025--3038.

\bibitem{dhara2020color}
S.~K. Dhara, M.~Roy, D.~Sen, P.~K. Biswas, Color cast dependent image dehazing via adaptive airlight refinement and non-linear color balancing, IEEE Trans. Circ. Syst. Vid. 31~(5) (2020) 2076--2081.

\bibitem{bartani2022adaptive}
A.~Bartani, A.~Abdollahpouri, M.~Ramezani, F.~A. Tab, An adaptive optic-physic based dust removal method using optimized air-light and transfer function, Multimedia Tools Appl. 81~(23) (2022) 33823--33849.

\bibitem{zhang2023underwater}
W.~Zhang, S.~Jin, P.~Zhuang, Z.~Liang, C.~Li, Underwater image enhancement via piecewise color correction and dual prior optimized contrast enhancement, IEEE Signal Process Lett. 30 (2023) 229--233.

\bibitem{lin2023smnet}
S.~Lin, F.~Tang, W.~Dong, X.~Pan, C.~Xu, Smnet: Synchronous multi-scale low light enhancement network with local and global concern, IEEE Trans. on Multimedia (Mar. 2023).

\end{thebibliography}

\end{document}